\pdfoutput=1
\documentclass[conference]{IEEEtran}
\IEEEoverridecommandlockouts
\usepackage{cite}
\usepackage{amsmath,amssymb,amsfonts}
\usepackage{algorithmic}
\usepackage{graphicx}
\usepackage{textcomp}
\usepackage{xcolor}
\usepackage{subfig}
\usepackage[super]{nth}
\usepackage[T1]{fontenc}
\usepackage{multirow}
\usepackage[margin=0.75in]{geometry}
\def\BibTeX{{\rm B\kern-.05em{\sc i\kern-.025em b}\kern-.08em
    T\kern-.1667em\lower.7ex\hbox{E}\kern-.125emX}}
\begin{document}

\addtolength{\topmargin}{0.25in}
\addtolength{\textheight}{-0.25in}

\title{Exploiting Vulnerability of Pooling in Convolutional Neural Networks by Strict Layer-Output Manipulation for Adversarial Attacks}

\author{Chenchen Zhao and Hao Li$^{*}$
\thanks{This research work is supported by the SJTU (Shanghai Jiao Tong Univ.) Young Talent Funding (WF220426002).}
\thanks{Chenchen Zhao is with Dept. Automation, SJTU, Shanghai, 200240, China. }
\thanks{Hao Li, Assoc. Prof., is with Dept. Automation and SPEIT, SJTU, Shanghai, 200240, China. }
\thanks{* Corresponding author: Hao Li ({Email:\tt\small haoli@sjtu.edu.cn})}
}

\maketitle

\begin{abstract}

Convolutional neural networks (CNN) have been more and more applied in mobile robotics such as intelligent vehicles. Security of CNNs in robotics applications is an important issue, for which potential adversarial attacks on CNNs are worth research. Pooling is a typical step of dimension reduction and information discarding in CNNs. Such information discarding may result in mis-deletion and mis-preservation of data features which largely influence the output of the network. This may aggravate the vulnerability of CNNs to adversarial attacks. In this paper, we conduct adversarial attacks on CNNs from the perspective of network structure by investigating and exploiting the vulnerability of pooling. First, a novel adversarial attack methodology named Strict Layer-Output Manipulation (SLOM) is proposed. Then an attack method based on Strict Pooling Manipulation (SPM) which is an instantiation of the SLOM spirit is designed to effectively realize both type I and type II adversarial attacks on a target CNN. Performances of attacks based on SPM at different depths are also investigated and compared. Moreover, performances of attack methods designed by instantiating the SLOM spirit with different operation layers of CNNs are compared. Experiment results reflect that pooling tends to be more vulnerable to adversarial attacks than other operations in CNNs.

\end{abstract}


\section{Introduction}\label{intro}

With the development of deep learning,
neural networks gradually dominate the field of pattern recognition tasks,
especially in complex scenes for which conventional vision methods may not be qualified.
For instance, convolutional neural networks (CNN) gradually take the lead
in those tasks, such as advanced robotics and autonomous driving
\cite{deepdriving,levine2016end,johns2014generative,kitti}.

However, although with proved high performance,
CNNs are also proved vulnerable to well-designed adversarial attacks,
as reported in recent studies \cite{1412_6572,1709_02802,1712_07107}.
Especially in fields stated above requiring high precision and have extremely low fault tolerance,
adversarial attacks on CNNs in related tasks may result in severe accidents and losses beyond measure.
For instance, a CNN based surgery machine may misclassify healthy tissues as tumors
and deliver the deadly order of excision;
a deep autopilot system may misrecognize pedestrians as background and fail to do obstacle avoidance.
Moreover, such approaches may turn malicious,
dealing heavy destructions in malignant competitions, crimes and terrorist activities.
Therefore, researchers have paid considerable attention to such attacks.

Feasibility of such adversarial attacks are mainly due to two characteristics of the CNN:
oversensitivity to non-core properties, and insensitivity to core properties of the data.
The core and non-core properties are represented by their influences on the class and features of the data,
defined by an oracle which has 100\% correct interpretation at all time.

Corresponding to the two characteristics,
data modification within a small threshold can result in different interpretations by the target network
(named type II attacks \cite{1412_6572,1608_04644,1709_04114,1706_06083,1805_07894}),
whereas data modification at a large scale can result in almost the same interpretation by the target network
(named type I attacks \cite{type1,1412_1897}).
Common type II attacks correspond to errors from sensors and external environment,
such as images captured by tainted cameras or in rainy days;
common type I attacks correspond to internal errors, such as false positives from scenes and objects.
According to literature on adversarial attacks and defences,
type II attacks have attracted more research interest than type I attacks by so far,
yet the two types of attacks are equally important
in analysing vulnerability of neural networks and in robust training.

In this paper, we investigate and exploit vulnerability of pooling
to conduct adversarial attacks on CNNs from the perspective of network structure.
Attacks in two frequently-neglected aspects are analyzed and conducted:
\begin{itemize}
\item Structure vulnerability of network layers.
Instead of digging into the training process and loss functions of the network model,
which are targeted by most attack methods,
the method proposed in this paper focuses on vulnerability of specific layers in the network.
Attacks of this kind are more fatal than existing attacks on account of the following reasons:
Firstly, the essential structure vulnerability of the layers is difficult or even impossible to eliminate;
secondly, such vulnerability may occur in core layers in networks
which cannot be arbitrarily replaced or modified;
thirdly, the vulnerable layers may be shared by multiple network models
which are all vulnerable to related attacks.
The proposed method offers a new perspective of possibly more severe attacks.
\item Type I scenarios.
Although frequently neglected, type I attacks are equally important as type II attacks
in network analyses and robustness studies,
since either of them may result in deadly judgement errors by the network model.
The proposed method is intended to be able to conduct both types of attacks, and is more comprehensive.
\end{itemize}

Our contributions are summarized as follows:
First, a novel adversarial attack methodology named Strict Layer-Output Manipulation (SLOM) is proposed.
Then an attack method based on Strict Pooling Manipulation (SPM) which is an instantiation of the SLOM spirit
is designed to effectively realize both type I and type II adversarial attacks on a target CNN.
Performances of attacks based on SPM at different depths are also investigated.
Moreover, a comparative study is carried out for attack methods designed by
instantiating the SLOM spirit with different operation layers of CNNs.
This comparative study reflects from a specific perspective that
pooling tends to be potentially more vulnerable to adversarial attacks,
compared with other operations such as convolution and activation of CNNs.

The paper is organized as follows:
Related works on adversarial attacks are stated in Section \ref{relatedwork}.
Detailed presentation of SLOM and implementations of SPM and SPM based attacks
are presented in Section \ref{core}.
Analyses on vulnerability of pooling are stated in Section \ref{theory}.
In Section \ref{experiment}, experiments are conducted to verify the statements in
Section \ref{core} and \ref{theory}, and investigate deeper-level characteristics of vulnerability of pooling.
We conclude the paper in Section \ref{conclusion}.

\addtolength{\topmargin}{-0.25in}
\addtolength{\textheight}{0.25in}

\section{Related Work}\label{relatedwork}

Although type I and type II attacks seem to have close relationship,
it is analyzed in \cite{type1} that type I and type II attacks are essentially different
and cannot be simply regarded as mutually equivalent variants of each other. 

Type II attacks are well-investigated and various high-performance attack methods are reported.
The authors in \cite{1312_6199} considered the type II attack as an optimization problem of perturbation
and solved it using the L-BFGS algorithm.
In \cite{1412_6572}, the authors proposed the Fast Gradient Sign Method (FGSM)
to maximize the output change caused by small perturbation. 
This milestone research work inspired many following algorithms \cite{1607_02533,1706_06083}.
An algorithm which aims at determining the minimum amount of perturbation
for a valid attack is proposed in \cite{1511_04599}.
In \cite{perturbation2019,perturbation2020}, the author proposed the feature purturbation method,
which is to compare the difference of the source and the target image on the activation of a specific layer.
Other type II attacks based on conditional generative models
(e.g. supervised VAE, conditional GAN)
\cite{type1,1411_1784,1406_5298,1511_05644},
box constraint algorithms \cite{1608_04644}
and input-output saliency mapping \cite{1511_07528} are also studied.

In contrast, type I attacks have attracted much less attention.
The authors in \cite{type1} used VAE to project the image to latent space,
and turned the perspective of gradient propagation from pixel-level to feature-level.
As presented in \cite{1412_1897},
specific evolutionary algorithms are used to select the most suitable adversarial examples.

For existing attack methods dedicated to the commonly-used CNN,
the vulnerability of each individual operation layer of the CNN is rarely studied.
Especially, the pooling process, which involves data dimension reduction
and may incur mistakes in both information preservation and discarding,
tends to be a vulnerable factor to both type I and type II attacks.
However, study on vulnerability of pooling as well as other important operations of the CNN is lack.
This paper puts forward some study on this non-negligible aspect.

\section{Strict Layer-Output Manipulation (SLOM) and SLOM-based Adversarial Attacks}\label{core}

\subsection{Methodology of SLOM}\label{slom}

SLOM is a methodology that focuses on using certain specially-generated data sample
to strictly manipulate the output of a specific layer of the target network and
aims at minimizing the difference between the layer output and a pre-defined label. 
The label is the output of the same layer of the target network
fed with the original data sample in type I attacks,
and the output fed with another data sample in type II attacks.
Compared with typical adversarial attack methods,
SLOM has a different criterion concerned with latent space features.
The difference is shown in \eqref{latentdiff1} in type I attacks
and \eqref{latentdiff2} in type II attacks,
in which $\overline{x}$ is the initial value of the adversarial data $x$;
$\hat{x}$ is a data sample of a class different from $\overline{x}$;
$f_{encoder}$ is composed of all operations before the target layer of the target network;
$J_1$ and $J_2$ are respectively mean squared error (MSE) adopted by SLOM
and cross entropy loss adopted by typical adversarial attack methods;
$f_{oracle}$ is the oracle stated in Section \ref{intro}.
\begin{equation}
\begin{aligned}
J_{SLOM}(x) & = J_1(f_{encoder}(x), f_{encoder}(\overline{x})) \\
J_0(x) & = J_2(f_{oracle}(x), f_{oracle}(\overline{x}))
\end{aligned}
\label{latentdiff1}
\end{equation}
\begin{equation}
\begin{aligned}
J_{SLOM}(x) & = J_1(f_{encoder}(x), f_{encoder}(\hat{x})) \\
J_0(x) & = J_2(f_{oracle}(x), f_{oracle}(\hat{x}))
\end{aligned}
\label{latentdiff2}
\end{equation}

Compared with feature perturbation, SLOM has the following characteristics:
\begin{itemize}
\item
SLOM is capable of realizing not only type II attacks for which feature perturbation methods are intended,
but also type I attacks which are beyond the capability of these related methods in the same field.
SLOM makes feature perturbation also qualified for the common and significant type I scenarios.
\item
SLOM involves analyses on convolution, activation and pooling separately
to generate multiple attack results on different layers.
Based on this, we investigate the vulnerability of the modules in CNN
and perform a comparative study among them.
The study reveals and exploits the potential vulnerability of pooling from a specific perspective.
This conclusion acts not only as a suggestion
about the choice of the target layer in existing feature perturbation methods,
but also as a reference for subsequent adversarial attack methods
about the specific structure breakpoint that they can target.
SLOM may serve as a general methodology guidance for realizing attacks
rather than purely a concrete attack technique.
\item
SLOM and existing feature perturbation methods have different principles of iteration and metrics,
meaning none of them can be used as baseline for SLOM.
Details of this are stated in Section \ref{experiment}.
\end{itemize}

In both type I and type II attacks,
SLOM has more strict restrictions on data output than typical adversarial attack methods.
The effectiveness of SLOM based attacks (as will be demonstrated afterwards) implies that
type I and type II attacks can be implemented not only by manipulation of classification results,
but also by more strict manipulation of intermediate outputs of layers in the target network.

\subsection{SPM Based Type I Attack}\label{t1attack}

SPM is an instantiation of the SLOM spirit. So are Strict Activation Manipulation (SAM) and
Strict Convolution Manipulation (SCM) stated in Section \ref{choice}.
SPM strictly manipulates the output of the target pooling layer.

\begin{figure}[tb]
\centering
\includegraphics[width=0.8\linewidth]{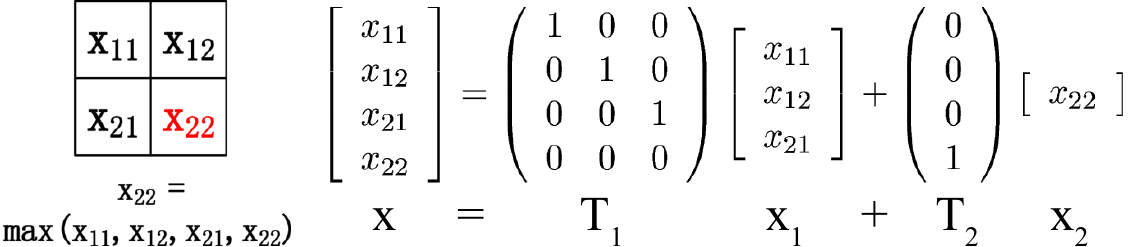}
\caption{Exemplary data decoupling and the corresponding mathematical representation in pooling}
\label{poolflatten}
\end{figure}

By decoupling the data according to the way shown in Fig. \ref{poolflatten},
the type I optimization can be represented by \eqref{pooloptim},
where $x_1$ and $x_2$ denote the non-maximum and maximum elements in the $n \times n$ patch $x$ of the data,
and $T_1$ and $T_2$ are the corresponding transform matrices.
\begin{equation}
\begin{aligned}
& Pooling(x) = x_2 \\
& J_P(x_1, x_2) = T_1 \cdot \sum_{x_1}max(x_1 - x_2, 0) + T_2 \cdot |x_2 - \overline{x_2}| \\
\end{aligned}
\label{pooloptim}
\end{equation}

In type I attacks, we aim at generating an adversarial data sample
(i.e. a data sample with considerable changes compared with the original one)
which has the same intermediate output at a target pooling layer as the original data sample does.
Since pooling preserves the maximum value $x_2$ in each patch, we try to generate a data sample
with $x_2 = \overline{x_2}$ in \eqref{pooloptim} at the target pooling layer.
In the meantime, $x_1 < x_2$ in the generated data should also be guaranteed.

As such kind of adversarial data sample and the original one
have exactly the same intermediate output at the target pooling layer,
we can naturally conclude that they also have exactly the same intermediate and final output
at all layers behind it,
which implies that an effective type I attack is performed on the target network.
The large feasible region of $x_1$ shows that SPM based type I attacks
take advantage of insensitivity of the target pooling layer to data modifications.

Following the type I attack paradigm stated in \cite{type1},
the optimization criterion for guiding type I adversarial data generation is given in \eqref{t1loss},
where $f_{encoder}$ is composed of all operations before the target pooling layer in SPM,
$\hat{y}$ is a pre-defined label of a class different from $\overline{x}$,
and $\lambda > 0$ is the equilibrium parameter.
\begin{equation}
\begin{aligned}
J_{SPM-I}(x) = &J_1(f_{encoder}(x), f_{encoder}(\overline{x})) + \\
               &\lambda J_2(f_{oracle}(x), \hat{y})
\end{aligned}
\label{t1loss}
\end{equation}

Inspired by \cite{type1}, the value of $\lambda$ is large initially,
to ensure enough data modifications to be detected by the oracle.
During modification, $\lambda$ turns small gradually until strict manipulation of $f_{encoder}(x)$ is achieved.

\subsection{SPM Based Type II Attack}\label{t2attack}

Following the type II attack paradigm stated in \cite{type1},
the optimization criterion for guiding adversarial data generation is given in \eqref{t2loss},
where $\hat{P}$ is a pre-defined value in latent space of a class different from $f_{encoder}(\overline{x})$.
\begin{equation}
\begin{aligned}
J_{SPM-II}(x) = &J_1(f_{encoder}(x), \hat{P}) + \\
                &\lambda J_2(f_{oracle}(x), f_{oracle}(\overline{x}))
\end{aligned}
\label{t2loss}
\end{equation}

SPM based type II attacks take advantage of the oversensitivity of pooling output to data modifications.
The oversensitive characteristic is detailedly stated in Section \ref{theory}.

\subsection{Other SLOM Instantiations in Adversarial Attacks}\label{choice}

SLOM is a general methodology that can not only be instantiated with pooling layers
but also with other kinds of main operation layers in CNNs.
For example, the SLOM spirit can be instantiated with activation layers and convolution layers,
forming Strict Activation Manipulation (SAM) and Strict Convolution Manipulation (SCM) respectively.

For SAM, we aim at generating adversarial data to strictly manipulate the output
of a specific activation layer.
We use ReLU as an example to demonstrate the process of SAM.
Since ReLU is composed of two linear functions, it can be considered as a linear operation
by splitting the $x$ domain up to two parts and analyzing them separately.
By decoupling the data to positive and negative elements
in the similar way to that shown in Fig. \ref{poolflatten},
the optimization problem with ReLU can be represented by \eqref{reluoptim},
in which $x_-$ and $x_+$ are respectively the negative and positive elements of the data $x$,
$x_+$ and $\overline{x_+}$ are respectively the positive elements of the adversarial and original data,
and $T_-$ and $T_+$ are the corresponding transform matrices similar to those in Fig. \ref{poolflatten}.
\begin{equation}
\begin{aligned}
& x_- = min(x, 0) \\
& x_+ = max(x, 0) \\
& x = T_- \cdot x_- + T_+ \cdot x_+ \\
& ReLU(x) = x_+ \\
& J_{A}(x_-, x_+) = T_- \cdot\sum_{x_-}max(x_-, 0) + T_+ \cdot\sum_{x_+}|x_+ - \overline{x_+}| \\
\end{aligned}
\label{reluoptim}
\end{equation}

Since ReLU preserves positive elements of the data, in SAM, we try to generate a data sample
that guarantees the restriction $x_+ = \overline{x_+}$ at the target activation layer.
Similar to pooling, it should also be guaranteed that $x_- \leq 0$ in the generated data.

The optimization criteria for type I and type II adversarial data generation based on SAM are similar to
those given in \eqref{t1loss} and \eqref{t2loss},
where $f_{encoder}$ is composed of all operations before the target activation layer in SAM.

For SCM, we aim at generating adversarial data
to strictly manipulate the output of a specific convolution layer.
The optimization criteria for type I and type II adversarial data generation based on SCM are similar to
those given in \eqref{t1loss} and \eqref{t2loss},
where $f_{encoder}$ is composed of all operations before the target convolution layer in SCM.

A comparative study of the SPM, SAM and SCM based attack methods is presented in Section \ref{experiment}.

\section{Vulnerability of Pooling in CNNs}\label{theory}

\begin{figure}[tb]
\centering
\subfloat[Neglecting of pixel gradient in mean pooling]{
\includegraphics[width=0.4\linewidth]{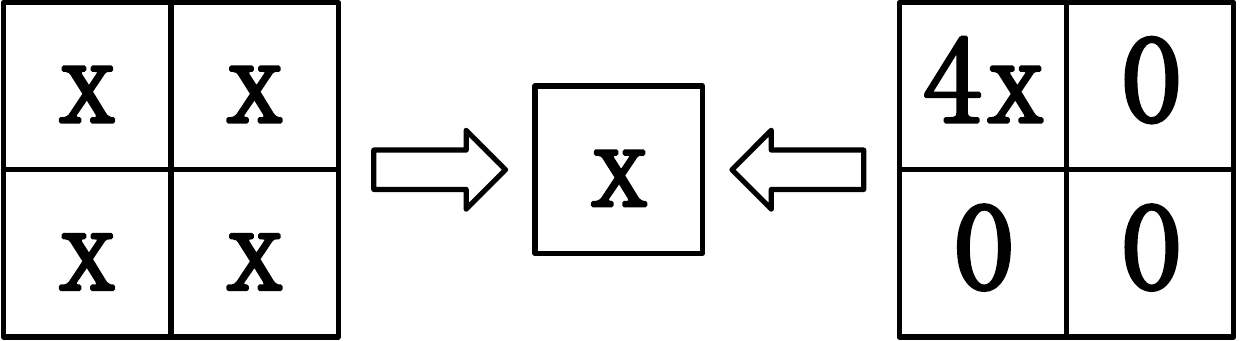}
}
\subfloat[Neglecting of similar-level features in max pooling]{
\includegraphics[width=0.4\linewidth]{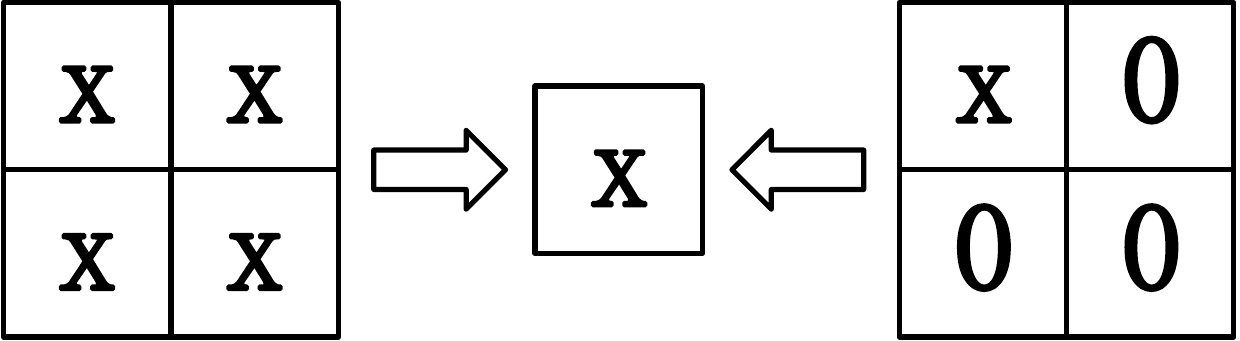}
}
\centering
\caption{Two exemplary situations of mis-preservation and mis-deletion of information caused by pooling}
\label{info-loss}
\end{figure}

There are mis-deletion and mis-preservation of data in pooling, even in well-trained CNNs.
Fig. \ref{info-loss} shows the above two situations of misbehaviour,
respectively caused by neglecting of pixel gradient and similar-level features of data.

Consider pooling as a linear matrix inequality (LMI) in max pooling:
each element in the pooling output is bigger than or equal to all the corresponding elements in the input;
or as a linear matrix equality (LME) in average pooling:
each element in the pooling output is the average of all the corresponding elements in the input.
Multiple data input may result in the same pooling output if the LMI/LME has multiple solutions,
i.e. the pooling operation is underconstrained.

A pooling operation with kernel size/stride $n$ after a convolution operation on data
with $c$ channels and width/height $a$ is underconstrained if \eqref{solution-pool} is satisfied,
in which $k$ and $m$ are the number and size of convolution kernels.
\begin{equation}
\begin{aligned}
& k < c \cdot n^2 \text{ (same padding)} \\
& k < c \cdot n^2 \cdot \frac{a^2}{(a-m+1)^2} \text{ (no padding)}
\end{aligned}
\label{solution-pool}
\end{equation}

According to \eqref{solution-pool}, with the popular $2 \times 2$ and stride 2 pooling,
the convolution and pooling operations become underconstrained
if the number of kernels is less than 4 times the number of channels of data.
This underconstrained condition frequently exists in popular CNN structures
such as VGG \cite{vgg} and ResNet \cite{resnet}, with the ratio usually 2 or 1.
When the underconstrained condition exists,
it is highly possible to find infinite number of data samples with exactly the same pooling output.
This implies that not only their classification results,
but also their probability distributions outputted by the target network are exactly the same.
Even if the underconstrained condition does not exist,
it is still highly possible to change the pooling output within a threshold
that do not lead to a change of its classification result,
which can be realized by general optimization methods.
Therefore, this underconstrained condition serves as theoretical basis for SPM based type I attacks.

With the same LME as the mathematical model, it can also be proved that pooling,
from this perspective, is more vulnerable than other modules in CNN.
A convolution operation on data with $c$ channels and width/height $a$ is underconstrained
if \eqref{solution-conv} is satisfied,
in which $k$ and $m$ are the number and size of convolution kernels.
\begin{equation}
\begin{aligned}
& k < c \text{ (same padding)} \\
& k < c \cdot \frac{a^2}{(a-m+1)^2} \text{ (no padding)}
\end{aligned}
\label{solution-conv}
\end{equation}

In typical CNNs with same-padding convolution layers, $k$ is bigger than or equal to $c$ (with ratio 2 or 1).
Therefore, convolutions in CNNs are mostly overconstrained,
and it is hard or even impossible to find multiple data samples with the same convolution output.
This means that SCM is much harder to conduct, and is more likely to fail than SPM.
On the other hand, activation is also a tougher target of SLOM than pooling.
Among all common activation functions, only ReLU is a many-to-one mapping.
However, the high fluctuation of the number of negative elements in the input data
adds to the unstability of SAM attacks.
In conclusion, among all instantiations of SLOM,
SPM has theoretically better performance and higher chance of success,
indicating that pooling is more vulnerable than convolution and activation from the SLOM perspective.

Besides, according to characteristics of convolution,
a slight change of data may cause a huge change of its pooling output,
and lead to change of the final classification result by the target network.
Consider one convolution operation on a $m \times m$ patch with kernel $kernel_{m \times m}$.
The relationship between a gaussian perturbation $\varepsilon_{data} \sim N(0, \sigma)$ on data
and the corresponding convolution output is as \eqref{convamp},
in which $\varepsilon_{output}$ is the perturbation caused by $\varepsilon_{data}$;
$K$ is the maximum absolute value of all elements in $kernel$.
\begin{equation}
\begin{aligned}
& \varepsilon_{output} \sim N(0, m^2 \cdot K \cdot \sigma) \\
& K = max(|kernel_{m \times m}|) \\
\end{aligned}
\label{convamp}
\end{equation}

Considering error accumulation,
the relationship between a gaussian perturbation $\varepsilon_{input} \sim N(0, \sigma)$ on input
and the final convolution output is as \eqref{convamp2},
in which $I$ is the number of convolution layers.
\begin{equation}
\begin{aligned}
& \varepsilon_{output} \sim N(0, \prod_i^I {m_i}^2 K_i \cdot \sigma) \\
& K_i = max(|{kernel}_i|) \\
\end{aligned}
\label{convamp2}
\end{equation}
Therefore, convolution amplifies the degree of data perturbation exponentially
given the condition $K_i > \frac{1}{{m_i}^2}$, which is a common case in CNNs.

As a consequence, SPM can also be used for type II attacks.
Note that with the overconstrained condition of convolution,
it is still hard to conduct SLOM on convolution layers even though they may amplify data perturbations.
It is pooling that makes the amplification available for type II attacks.

\section{Experiments and Results}\label{experiment}

\subsection{Experiment Setup}\label{setup}

We conduct experiments with MNIST (grayscale) \cite{mnist},
ImageNet (colored) \cite{imagenet} and KITTI \cite{kitti} datasets.
The oracle is represented by 5 identical classifiers with independent training processes.
For attacks with ImageNet and KITTI,
we adopt the widely-used VGG16 \cite{vgg} and ResNet18 \cite{resnet} networks;
for attacks with MNIST, we construct simpler versions of CNN as the target and the oracle networks.
Configurations of the CNNs in MNIST experiments are shown in Table \ref{netconf}.
\begin{table}[tb]
\centering
\begin{tabular}{|c|c|}
\hline
Target & Oracle \\
\hline
\multirow{2}*{conv3-4} & conv3-16 \\
\cline{2-2}
 & conv3-32 \\
\hline
\multicolumn{2}{|c|}{maxpooling-2} \\
\hline
\multirow{2}*{conv3-8} & conv3-64 \\
\cline{2-2}
 & conv3-64 \\
\hline
\multicolumn{2}{|c|}{maxpooling-2} \\
\hline
\multicolumn{2}{|c|}{FC-1024} \\
\hline
\multicolumn{2}{|c|}{FC-10} \\
\hline
\end{tabular}
\caption{CNN configurations of the target and the oracle networks in MNIST experiments}
\label{netconf}
\end{table}

\begin{figure}[tb]
\centering
\subfloat[Type I examples.]{
\includegraphics[width=0.45\linewidth]{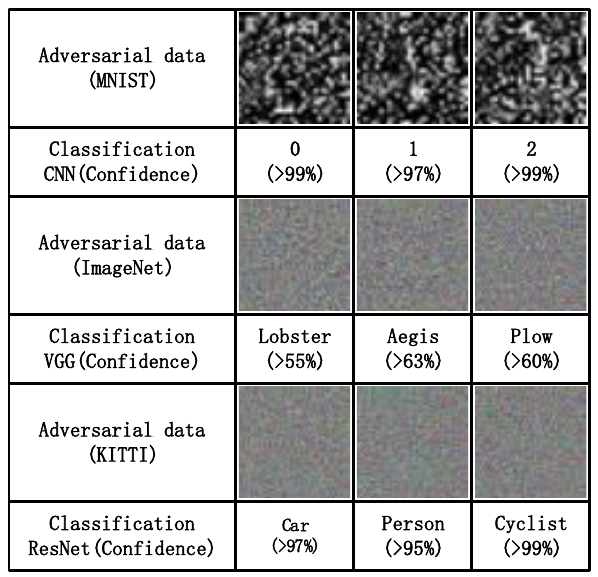}
\label{t1data}
}
\subfloat[Type II examples.]{
\includegraphics[width=0.45\linewidth]{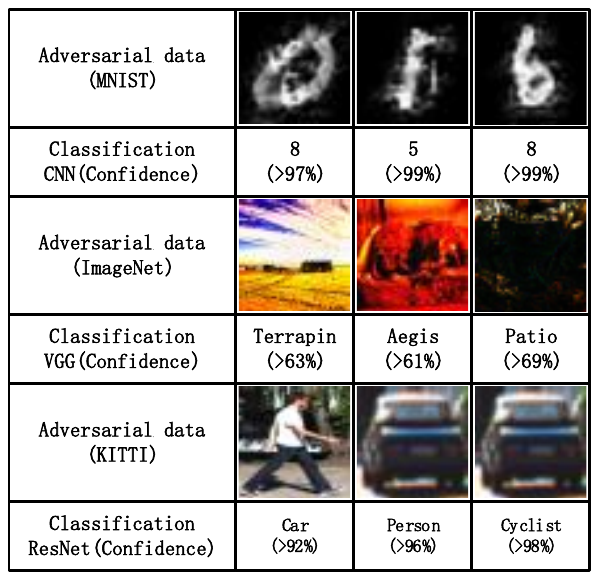}
\label{t2data}
}
\caption{Adversarial examples generated by SPM.}
\label{attack_data}
\end{figure}

For type I attacks, noise-like adversarial examples (`fooling' images \cite{1412_1897})
are generated to be classified into a pre-defined class by the target network with high confidence;
for type II attacks, data samples are generated to be classified into a different class
with high confidence but have very small amplitude of pixel-level changes.
Every data sample goes through 30K iterations.
For attacks with MNIST, iteration step length is set $10^{-3}$ in the first 10K iterations,
and $10^{-4}$ in the last 20K iterations;
for attacks with ImageNet and KITTI, iteration step length is set $10^{-4}$ in the first 10K iterations,
and $10^{-5}$ in the last 20K iterations.
Oracles are not adopted in any attack,
since they are a-posteriori-proved unnecessary by the oracle-free experiments.
Oracle-free experiments can also amplify the performance difference
between the three SLOM based attacks,
since it is easy to observe possible violations of type I and type II rules stated in \cite{type1}.

In the comparative experiments between SPM, SAM, and SCM,
the data goes through 800 iterations with step length 1.
In type I experiments, a noise-like adversarial example is generated to be classified as 0 by the network;
in type II experiments, a data sample with class 0 is modified to be classified as 3 by the network.

\subsection{Type I \& II Attacks and Results}\label{attack_result}

Fig. \ref{attack_data} shows the mis-classified adversarial data with high confidence by the target networks.
Fig. \ref{t1data} shows that
the criterion in \eqref{t1loss} has the ability to change noise-like images
to make their pooling outputs similar to ordinary ones while remaining their `noisy' characteristic.
Fig. \ref{t2data} shows that pooling bridges
the amplification of modifications by convolution and type II attacks.
In conclusion, experiment results show the effectiveness of SPM,
reflecting potential vulnerability of pooling to type I and type II attacks.

\begin{figure}
\centering
\includegraphics[width=0.85\linewidth]{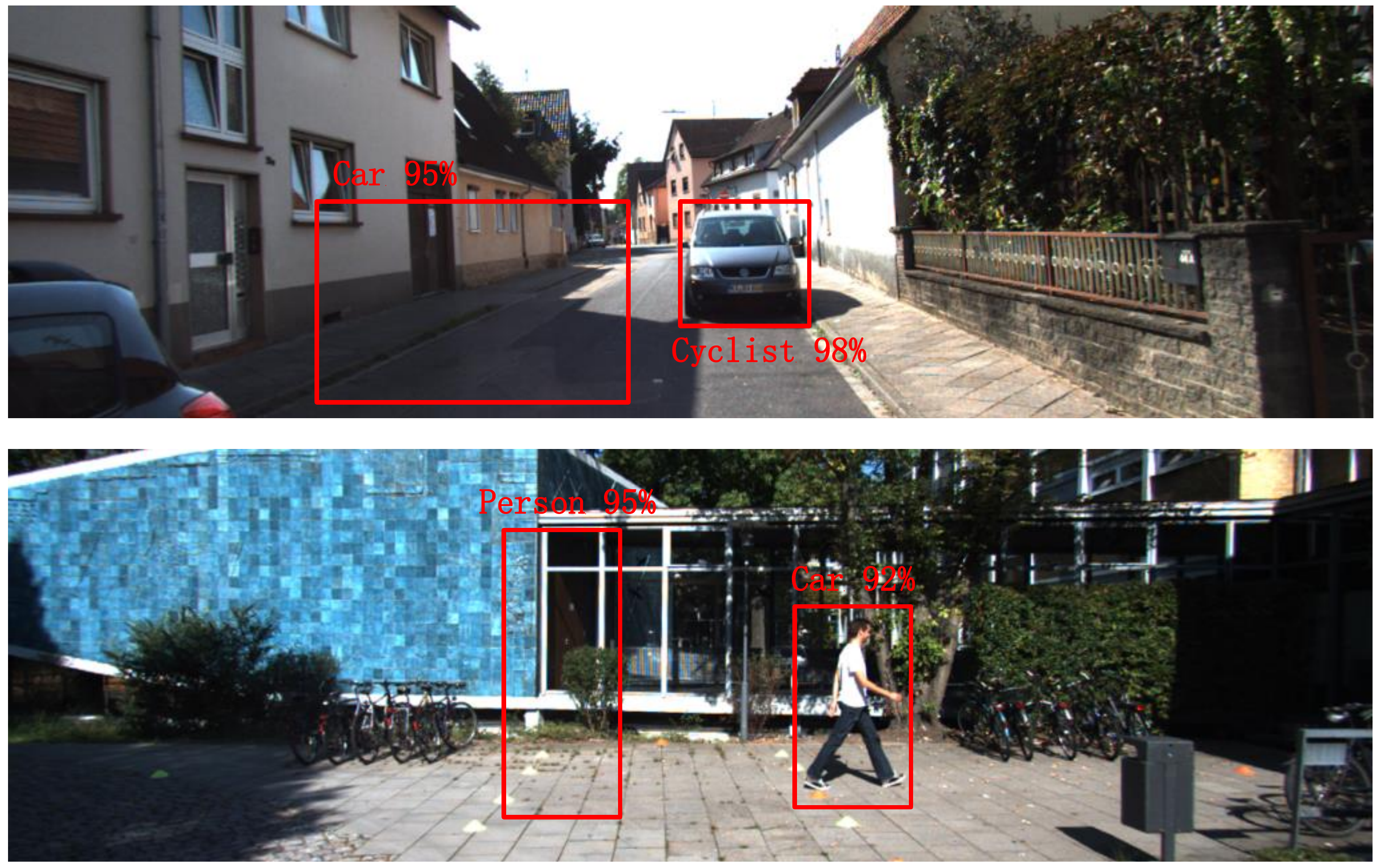}
\caption{The combinational SPM based attack confuses the object detection and recognition network.}
\label{combinedresult}
\end{figure}

By combining the SPM based type I and type II attacks,
the proposed method can severely confuse the network to output totally wrong results.
By attacking with selected patches from the original data,
the combinational attack confuses the network to fail the object detection and recognition task,
with the outputs shown in Fig. \ref{combinedresult}.

\subsection{Performance Comparison among SPM, SAM and SCM Based Attacks}\label{opcomparison}

The encoder of the target network has two convolution layers, two activation layers and two pooling layers,
with its structure shown in Table \ref{netconf}.
In this section, strict manipulations of outputs of the 6 layers are respectively experimented
with identical data sample and iteration process, with their attack performance compared.
\begin{figure}[tb]
\centering
\begin{minipage}{0.55\linewidth}
\subfloat[Data modifications]{
\includegraphics[width=0.95\linewidth]{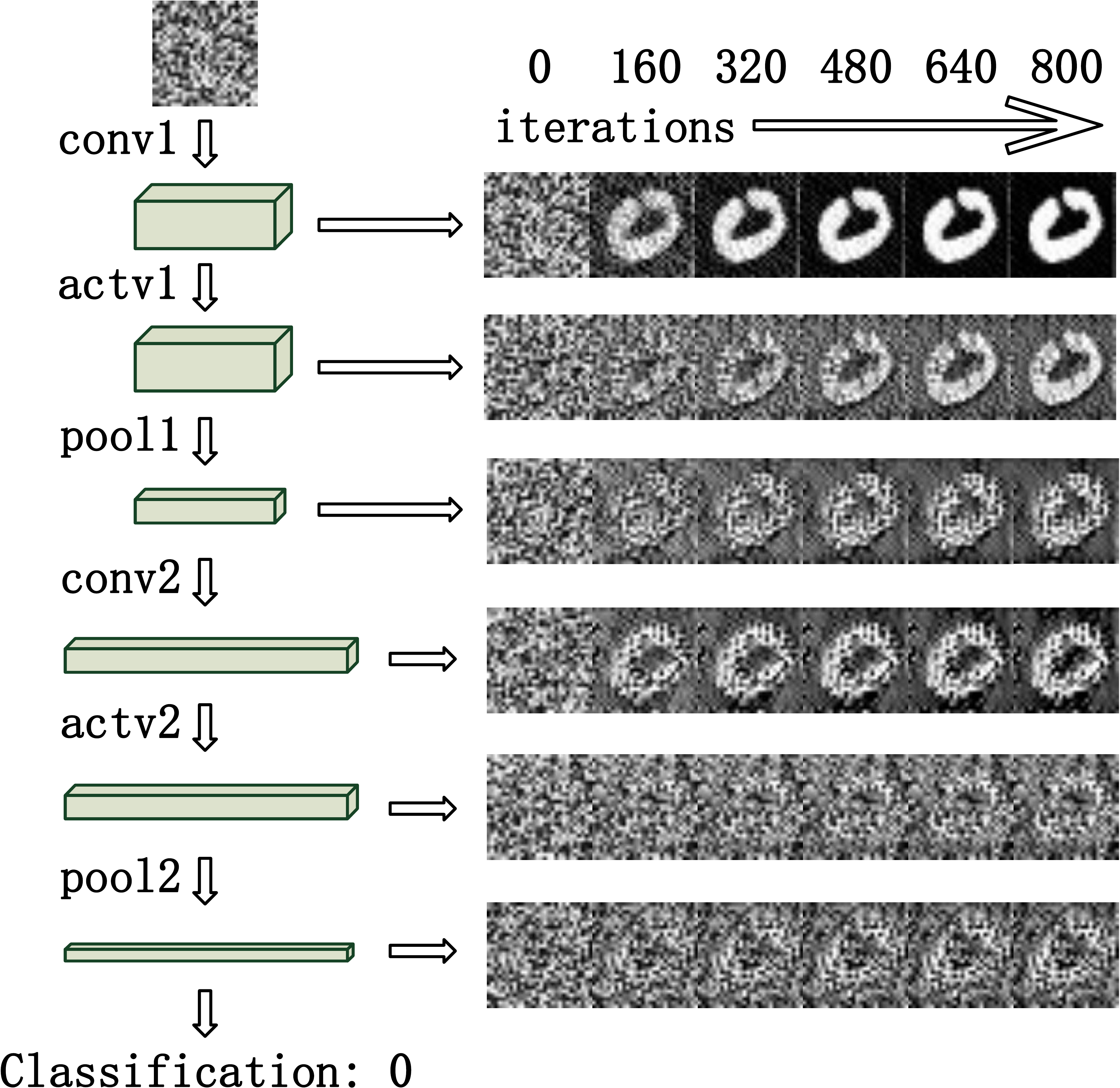}
\label{type1_process}
}
\end{minipage}
\begin{minipage}{0.35\linewidth}
\subfloat[Loss convergence]{
\includegraphics[width=0.95\linewidth]{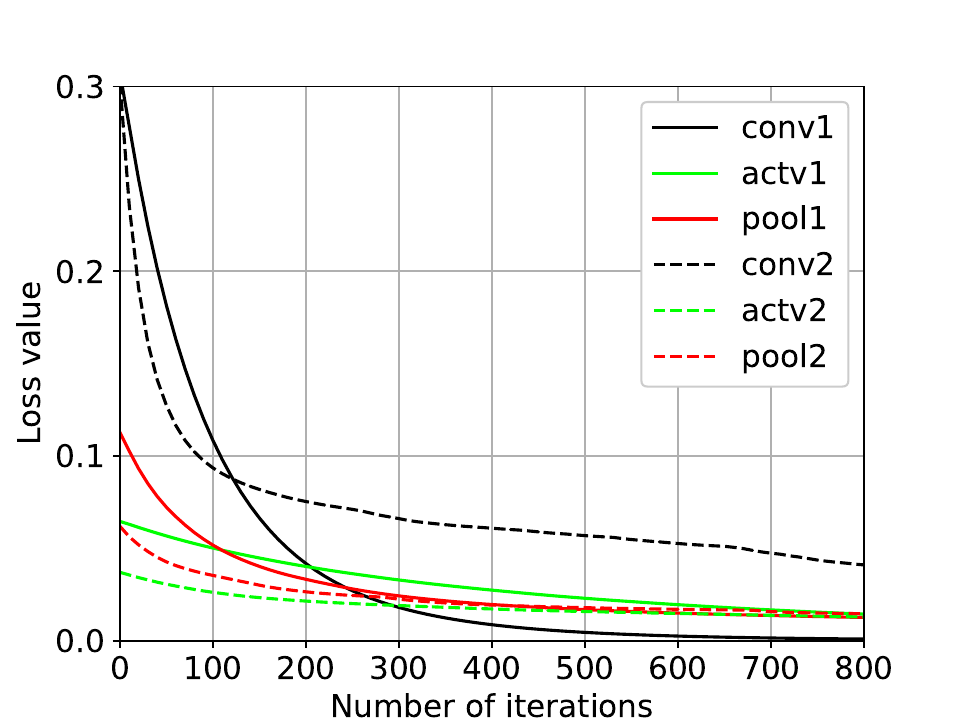}
\label{t1lossval}
}

\subfloat[Classification confidence]{
\includegraphics[width=0.95\linewidth]{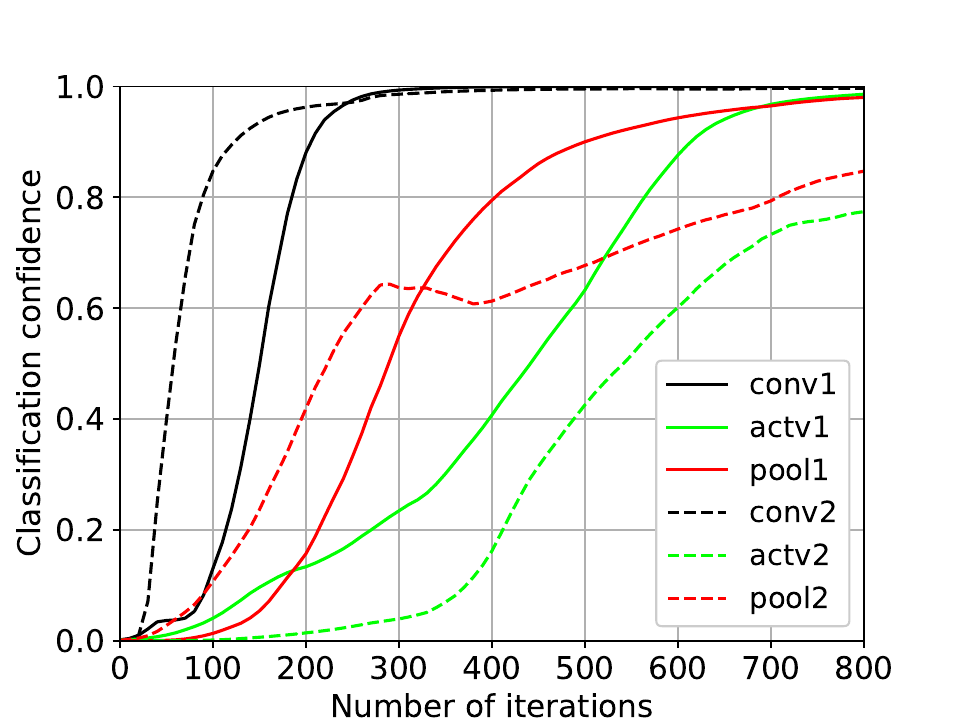}
\label{t1confidence}
}
\end{minipage}
\caption{Comparisons of SPM, SAM and SCM in type I attacks}
\label{t1comparison}
\end{figure}

Comparisons of SPM, SAM and SCM in type I attacks are shown in Fig. \ref{t1comparison}.
From Fig. \ref{type1_process},
it is obvious that in every block of network layers (conv + activation + pooling),
SCM has the modified data most similar to the original one,
indicating that it has the worst attack performance.
On the contrary, SPM has the highest level of attack performance.
With reference to the changing processes of loss and classification confidence of the target network
respectively shown in Fig. \ref{t1lossval} and Fig. \ref{t1confidence}, possible reasons are as follows:
\begin{itemize}
\item
Strong overconstraint of convolution:
convolution is the most overconstrained operation among the three main operations.
As a result, SCM has relatively the lowest fault tolerance,
meaning that the noise initialization will eventually be `perfectly' corrected
with very little difference from the original data.
This also accounts for the fact that convolution has the fastest loss convergence.
Fig. \ref{type1_process} indicates the correction process in SCM.
\item
The greatest dimension reduction in pooling: great dimension reduction in pooling
ensures the ability of SPM to tolerate high-amplitude data modifications.
Therefore, adversarial data generated by SPM has the best quality among the three instantiations of SLOM.
\end{itemize}
\begin{figure}[tb]
\centering
\begin{minipage}{0.55\linewidth}
\subfloat[Data modifications]{
\includegraphics[width=0.95\linewidth]{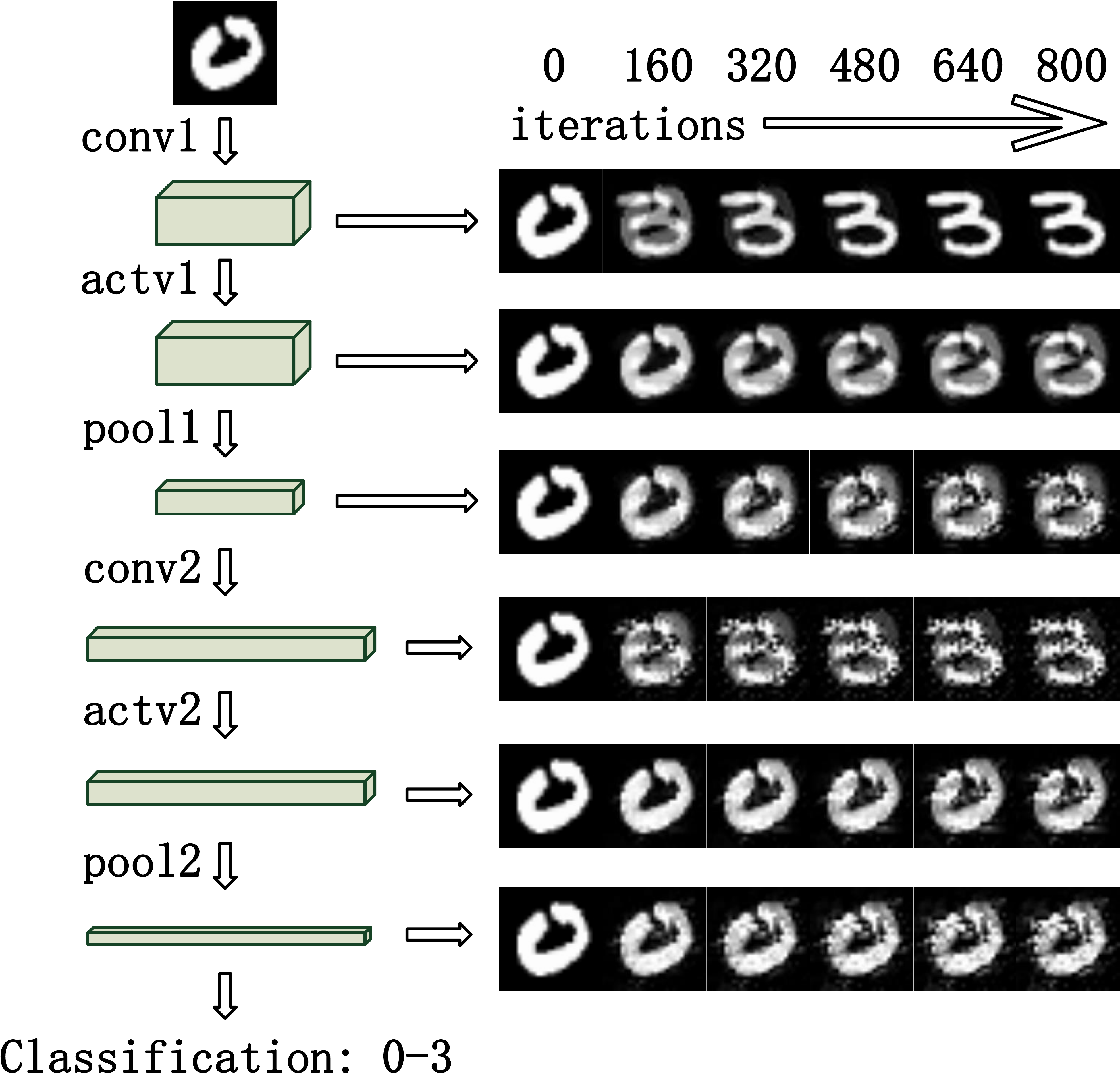}
\label{type2_process}
}
\end{minipage}
\begin{minipage}{0.35\linewidth}
\subfloat[Loss convergence]{
\includegraphics[width=0.95\linewidth]{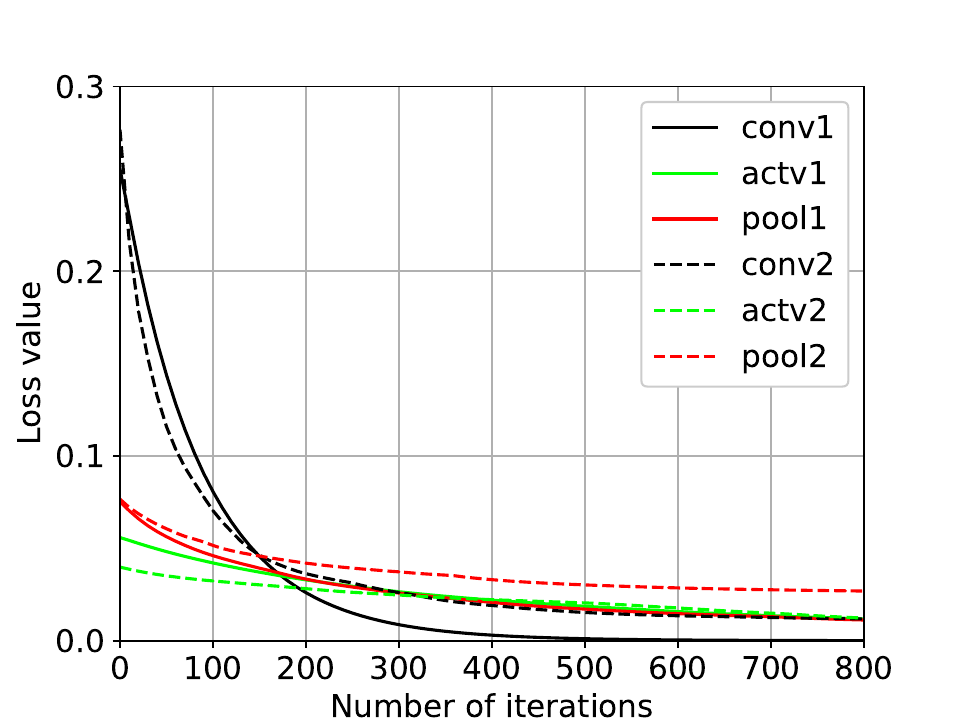}
\label{t2lossval}
}

\subfloat[Classification confidence]{
\includegraphics[width=0.95\linewidth]{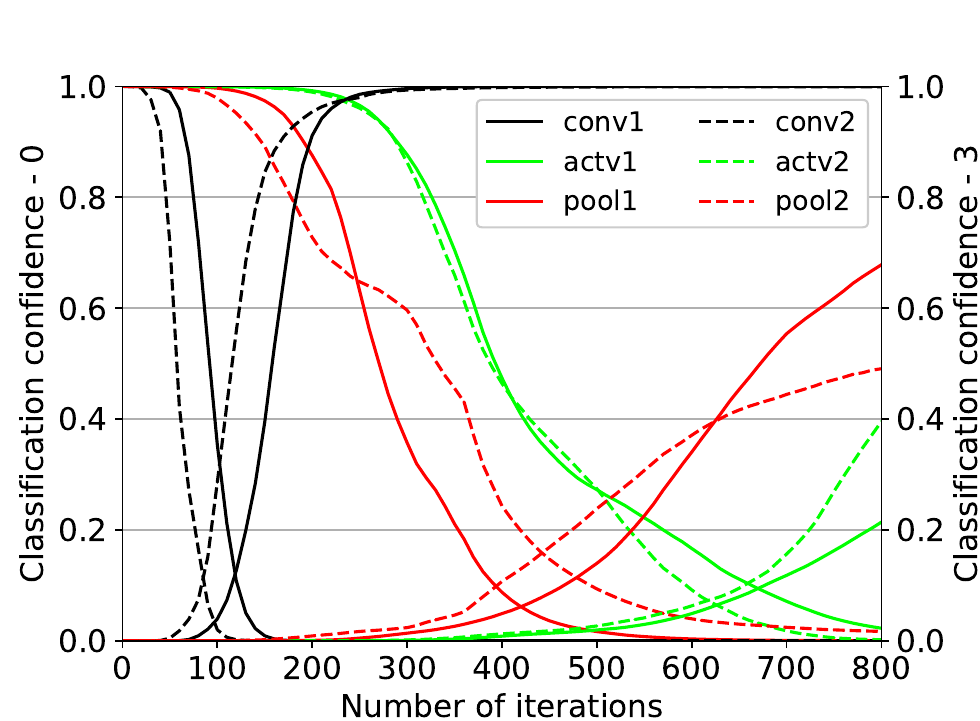}
\label{t2confidence}
}
\end{minipage}
\caption{Comparisons of SPM, SAM and SCM in type II attacks}
\label{t2comparison}
\end{figure}

Comparisons of SPM, SAM and SCM in type II attacks are shown in Fig. \ref{t2comparison}.
Similar to type I attacks, SPM has the best attack performance, while SCM has the worst.
Possible reasons are similar to those in type I attacks.
From the results in type II attacks,
although convolution may amplify the changes of data and reflect it in convolution outputs,
it is pooling that makes use of such amplified changes of output for type II attacks.
This proves the correctness of the theoretical analyses in Section \ref{theory}.

\subsection{Performance Comparison among SPM at Different Depths}\label{depthcomparison}

By comparing results with the two pooling layers, we may reach the following conclusions that
with the target pooling layer in SPM deeper,

\begin{itemize}
\item
Difference between the modified and the original ordinary data is bigger in type I attacks,
and smaller in type II attacks (i.e. attack performance increases in both types).
The deeper pooling layer has higher fault tolerance,
since every block in the target network corresponds to an underconstrained problem,
and the underconstrained characteristic can be accumulated.
\item
The loss convergence is slower.
Apart from the larger feasible region of data caused by higher fault tolerance,
the network becomes harder to train with the network structure deeper and more complex,
and the convergence process is certain to be slower.
\end{itemize}

\subsection{Discussion}\label{summary}

Experiment results show that among the three attack methods,
SPM is most effective in type I/II attacks.
SAM is barely qualified for both attacks,
but has much worse performance than SPM.
SCM is qualified for neither types of attacks.

Note that we do not conduct baseline comparisons on SLOM,
since there is yet no intuitive and objective shared metric of them.
One major difference between SLOM and other methods (including feature perturbation) is that
attack success rate cannot be directly applied to evaluation of SLOM.
With no oracle, SLOM modifies the original data
with much less constraints than other attack methods (including feature perturbation).
After enough iterations, the generated data is certain to have the expected classification results.
While SLOM seems to get almost 100\% success rate, the data may violate the type I and type II rules
stated in \cite{type1}.
Therefore, the metric of SLOM should instead be quality of the adversarial data.
On the contrary, success rate is the commonly-used metric in existing adversarial attack methods.
However, we have been investigating on the proper metric and experimental comparisons.
This will be an important improvement work in our future research.

Here, we do not imply that activation and convolution
which are not so affected by strict manipulation are resistant to adversarial attacks. 
Instead, the presented comparative study offers a new structural perspective,
with revelation and exploitation of potentially more severe vulnerability of pooling to adversarial attacks
compared with other two main operations of CNNs.

\section{Conclusion}\label{conclusion}

In this paper, we propose the adversarial attack methodology of Strict Layer-Output Manipulation (SLOM)
that uses certain specially-generated data sample
to strictly manipulate the output of a specific layer of the target network.
An adversarial type I and type II attack method named Strict Pooling Manipulation (SPM)
which is an instantiation of SLOM is proposed.
We conduct type I and type II experiments to show that
SPM is qualified for generation of type I and type II adversarial data.
We also demonstrate from the SLOM perspective that pooling has potentially the most severe vulnerability
among the three main operations in CNNs,
by a comparative study with SPM and SLOM with convolution/activation layers.
In conclusion, with proper manipulation of the related layer output,
vulnerability of pooling can be effectively exploited for adversarial data generation. 
The presented works will serve as a valuable guide for studying effective protection methods against CNN adversarial attacks in robotics applications.

\bibliographystyle{unsrt}
\bibliography{paper}

\end{document}